%% file: arxiv.tex
\documentclass[10pt,twocolumn,letterpaper]{article}

\input{macro}

\begin{document}

%%%%%%%%% TITLE - PLEASE UPDATE
\title{\TITLE}

\author{
Dave Zhenyu Chen$^{1}$ \quad Ronghang Hu$^{2}$ \quad Xinlei Chen$^{2}$ \quad Matthias Nie{\ss}ner$^{1}$ \quad Angel X. Chang$^{3}$\\
\qquad \\
$^{1}$Technical University of Munich \qquad $^{2}$Meta AI \qquad $^{3}$Simon Fraser University \\
% \institute{}
}

\input{figures/teaser}
%%%%%%%%% ABSTRACT
\begin{abstract}
    \input{sections/0_abstract}
\end{abstract}

%%%%%%%%% BODY TEXT

\input{sections/1_intro}
\input{sections/2_related_work}
\input{sections/3_method}
\input{sections/4_experiments}
\input{sections/5_conclusion}
\input{sections/7_acknowledgements}

%%%%%%%%% REFERENCES
{\small
\bibliography{egbib}
}

\input{sections/6_supplemental}

\end{document}

%% file: macro.tex
\usepackage[pagenumbers]{cvpr} % To force page numbers, e.g. for an arXiv version

\usepackage{graphicx}
\usepackage{amsmath}
\usepackage{amssymb}
\usepackage{booktabs}
\usepackage{multirow}
\usepackage{caption}
\usepackage{subcaption}
\usepackage{float}
\usepackage{subcaption}
\usepackage{enumitem}
\usepackage{numprint}
\usepackage{xcolor}
\usepackage{tabu}
\usepackage[numbers]{natbib}
\definecolor{GreenCanvas}{HTML}{F1F8EC}
\definecolor{RedCanvas}{HTML}{FCE9DC}

\usepackage[pagebackref,breaklinks,colorlinks]{hyperref}

\usepackage[capitalize]{cleveref}
\crefname{section}{Sec.}{Secs.}
\Crefname{section}{Section}{Sections}
\Crefname{table}{Table}{Tables}
\crefname{table}{Tab.}{Tabs.}

\def\cvprPaperID{6607} % *** Enter the CVPR Paper ID here
\def\confName{CVPR}
\def\confYear{2023}

\newcommand{\ARCH}{UniT3D\xspace}
\newcommand{\TITLE}{\ARCH: A Unified Transformer for 3D Dense Captioning and Visual Grounding}

\newcommand{\mypara}[1]{\noindent\textbf{#1}}

%% file: figures/teaser.tex
\twocolumn[{%
	\renewcommand\twocolumn[1][]{#1}%
	\maketitle
	\thispagestyle{empty}
	\begin{center}
		\includegraphics[width=0.99\textwidth]{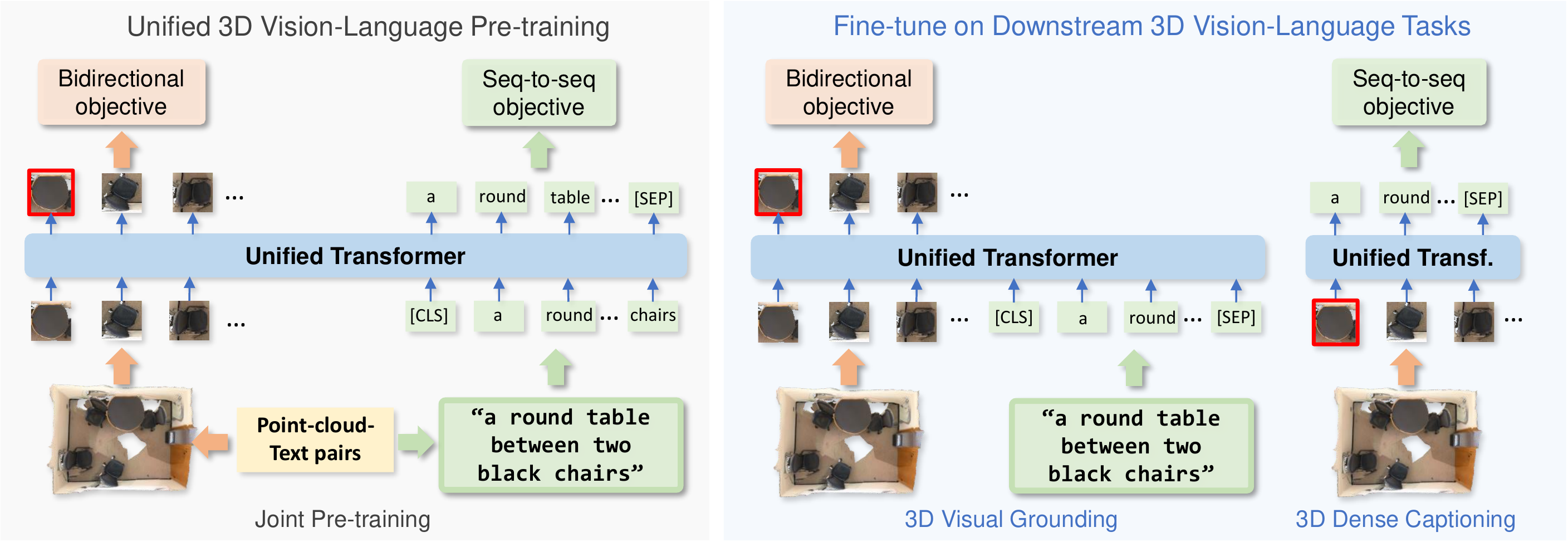}
		\captionof{figure}{
		We present~\ARCH, a unified transformer for 3D dense captioning and visual grounding.~\ARCH~is pre-trained with both bidirectional and seq-to-seq objectives on point-cloud-text pairs. Afterwards, it is further fine-tuned for 3D visual grounding and dense captioning. We show that our proposed architecture and pre-training scheme largely improve the performance on both downstream tasks.
		}
		\label{fig:teaser}
	\end{center}
}]

%% file: sections/0_abstract.tex
Performing 3D dense captioning and visual grounding requires a common and shared understanding of the underlying multimodal relationships.
However, despite some previous attempts on connecting these two related tasks with highly task-specific neural modules, it remains understudied how to explicitly depict their shared nature to learn them simultaneously.
In this work, we propose~\ARCH, a simple yet effective fully unified transformer-based architecture for jointly solving 3D visual grounding and dense captioning.
\ARCH enables learning a strong multimodal representation across the two tasks through a supervised joint pre-training scheme with bidirectional and seq-to-seq objectives.
With a generic architecture design,~\ARCH allows expanding the pre-training scope to more various training sources such as the synthesized data from 2D prior knowledge to benefit 3D vision-language tasks.
Extensive experiments and analysis demonstrate that~\ARCH obtains significant gains for 3D dense captioning and visual grounding.

%% file: sections/1_intro.tex
\section{Introduction}

The 3D vision-language field has been drawing increasing research interest in jointly understanding 3D scenes~\citep{dai20183dmv, qi2018frustum, lahoud20193d, qi2019deep, hou20193d, engelmann20203d, qi2020imvotenet, hou2020revealnet, cheng2021back} and natural language ~\citep{vaswani2017attention, devlin2018bert, liu2019roberta, yang2019xlnet, brown2020language}, such as 3D visual grounding~\citep{chen2020scanrefer} and 3D dense captioning~\citep{chen2021scan2cap}.
The task of 3D visual grounding takes as input a point-cloud-text pair and outputs a bounding box of the referred object. As its sibling task, 3D dense captioning expects a point cloud as input and densely generates object bounding boxes and descriptions in the scene.
Both tasks enable applications such as assistive robots and natural language control in AR/VR systems.

Although these two 3D vision-language tasks are naturally complementary to each other, previous attempts to connect them~\citep{chen2021d3net, cai20223djcg} only exploit partly shared object spatial information, leaving the joint nature between object relationships and textual semantics underdeveloped.
More concretely, to localize the target object in the scene, 3D visual grounding methods require a joint understanding of object attributes and spatial relationships. This requirement persists for 3D dense captioning when densely describing the appearance and spatial aspects of the objects.
Therefore, it is naturally desirable to develop a shared representation between two tasks with a unified task-agnostic framework, where the two input modalities are jointly encoded and enhanced.
Moreover, such generic multimodal representation is not only beneficial for sharing knowledge between visual grounding and dense captioning, but could also enable 3D vision-language research beyond specific domains, as in the case of VisualBERT~\citep{li2019visualbert} or CLIP~\citep{radford2021clip} on 2D images.

To this end, we propose UniT3D, a joint transformer-based solution to facilitate vision-language representation learning for 3D visual grounding and dense captioning, as illustrated in Fig.~\ref{fig:teaser}. Unlike prior works with two distinct and task-specific neural modules~\citep{cai20223djcg, chen2021d3net}, our method tackles the tasks of 3D dense captioning and visual grounding via a task-agnostic unified transformer with light-weight output heads. 
To enable joint vision-language representation learning, we design a supervised training scheme that combines the bidirectional objective with query-aware object matching supervision and the seq-to-seq objective with object-aware sequence generation supervision.
Through fine-tuning on specific tasks with corresponding output heads, the joint vision-language representation learned by our task-agnostic multimodal transformer is well capable of supporting both the localization of the referred objects in the scenes and the dense generation of object descriptions. 

One challenge of the joint 3D vision-language representation learning is that existing 3D vision-language datasets~\citep{chen2020scanrefer, achlioptas2020referit3d} are relatively limited in size and variety compared to their 2D counterparts such as MSCOCO~\citep{chen2015microsoft} and Conceptual Captions~\citep{sharma2018conceptual}. To address this challenge, we build a large-scale 3D vision-language dataset with text annotations generated from an image captioner learned on abundant 2D image and text datasets.
Concretely, we apply the image captioner from~\citet{mokady2021clipcap} to ScanNet images to obtain synthetic image-text pairs and convert them to point-cloud-text pairs by cropping the reconstructed point clouds in ScanNet within the image frustum using camera parameters as our synthetic 3D vision-language data.
We show that along with jointly pre-training on such synthesized data with the proposed bidirectional and seq-to-seq objectives, our UniT3D model obtains significant performance gains on the downstream 3D vision-language tasks.
To summarize, our contributions are threefold:

\begin{itemize}
\item We introduce a multimodal transformer architecture to solve 3D visual grounding and dense captioning in a fully unified fashion.
\item We propose a supervised joint pre-training scheme with bidirectional and seq-to-seq objectives to facilitate multimodal feature learning.
\item We construct a large-scale synthetic point-cloud-text dataset, showing that the distilled 2D prior knowledge is beneficial to 3D vision-language tasks.
\end{itemize}

%% file: sections/2_related_work.tex
\section{Related work}

\mypara{3D visual grounding and dense captioning.}
Recently, there has been a thriving research interest in 3D vision-language~\citep{chen2020scanrefer, achlioptas2020referit3d}. Seminal works~\citep{chen2020scanrefer, achlioptas2020referit3d} concurrently propose ScanRefer and ReferIt3D dataset consisting of free-form descriptions of 3D real-world object from ScanNet~\citep{dai2017scannet} scenes.
~\citet{chen2020scanrefer} propose the 3D visual grounding task to localize a target object in a 3D environment using text queries.
The majority of the previous work on 3D visual grounding~\citep{chen2020scanrefer, achlioptas2020referit3d, huang2021text, yuan2021instancerefer, zhao20213dvg, cai20223djcg, chen2021d3net} concentrate on distinguishing the multimodal relationships among objects and text queries.
As its reversed task, the task of 3D dense captioning predicts the bounding boxes and the associated descriptions for all objects in the input 3D scene~\citep{chen2021scan2cap}. 
Similar to 3D visual grounding, most prior works in this track ~\citep{chen2021scan2cap, yuan2022x, jiao2022more} learn the multimodal relationships among the objects in the 3D scene and decode them as text outputs.

We observe that these two tasks are complementary in nature, with minor discrepancies in their outputs (boxes vs. descriptions). There are several initial attempts in bridging them together.~\citet{chen2021d3net} apply a speaker-listener architecture to self-critically improve the overall performance of visual grounding and dense captioning. However, there is no shared representation between the visual and text modalities.~\citet{cai20223djcg} link both tasks with a shared detection backbone attached to two task-specific neural modules. Despite the shared object relationships within the visual modality, the multimodal representations between objects and text inputs are not explicitly modeled.
In contrast, our method directly builds multimodal representations with a joint multimodal fusion transformer, enabling both visual grounding and dense captioning prediction from shared multimodal knowledge. 

\input{figures/fine-tuning}

\mypara{Vision-language pre-training.}
In light of the shared nature of vision-language tasks, developing a joint multimodal representation between the vision and language modalities has become a popular research domain in the 2D vision-language community~\citep{li2019visualbert, lu2019vilbert, zhou2020unified, wang2021simvlm, hu2021unit, singh2022flava, wang2022ofa}. 
Inspired by the masked language modeling in~\citet{devlin2018bert}, seminal works such as VisualBERT~\citep{li2019visualbert} and ViLBERT~\citep{lu2019vilbert} introduce the masked vision-language modeling scheme for learning the multimodal representation. However, such pre-training schemes only concentrate on multimodal encoding, but neglect multimodal decoding, which is essential for generation-based tasks (e.g. image captioning). Based on masked vision-language modeling,~\citet{zhou2020unified} design the bidirectional and seq-to-seq pre-training objectives to empower both multimodal encoding and decoding.~\citet{hu2021unit} further expand the spectrum of the applicable downstream tasks using the learned multimodal representation. 
However, due to the limited amount of existing vision-language data in 3D, such pre-training strategies entangled with large transformer architectures cannot be directly migrated to the 3D vision-language domain. Hence, We propose a synthetic vision-language data generation scheme featuring the distilled 2D multimodal knowledge to facilitate the multimodal pre-training in the 3D domain.

\mypara{3D vision with 2D priors.}
Leveraging 2D distilled knowledge such as CLIP~\citep{radford2021clip} for 3D vision is a trending research topic. 
PointCLIP~\citep{zhang2022pointclip} transfers 2D knowledge to 3D by conducting alignments between the CLIP-encoded point cloud image and category texts. This scheme enables zero-shot classification on point cloud without training on any 3D datasets such as ShapeNet~\citep{chang2015shapenet}. 
CLIP2Point~\citep{huang2022clip2point} aligns depth image features to CLIP-encoded point cloud image features to facilitate image-depth pre-training for 3D shape understanding. SemAbs~\citep{ha2022semantic} extracts relevancy maps from CLIP for open-vocabulary 3D scene understanding and CLIP-NeRF~\citep{wang2022clip} leverages CLIP embedding space for 3D scene manipulation. 
More advanced methods utilize pixel-to-point correspondence~\citep{liu20213ddistillation, hamdi2021voint} or kernel inflation technique~\citep{xu2021image2point} to tackle 3D shape/scene understanding tasks leveraging 2D distilled knowledge.
In this work, we propose to use 2D vision-language priors to address the limited scale and variety of existing 3D data, enabling multimodal representation learning for 3D vision-language.

%% file: figures/fine-tuning.tex
\begin{figure*}[!ht]
    \centering
    \begin{subfigure}[b]{0.5\textwidth}
         \centering
         \includegraphics[width=\textwidth]{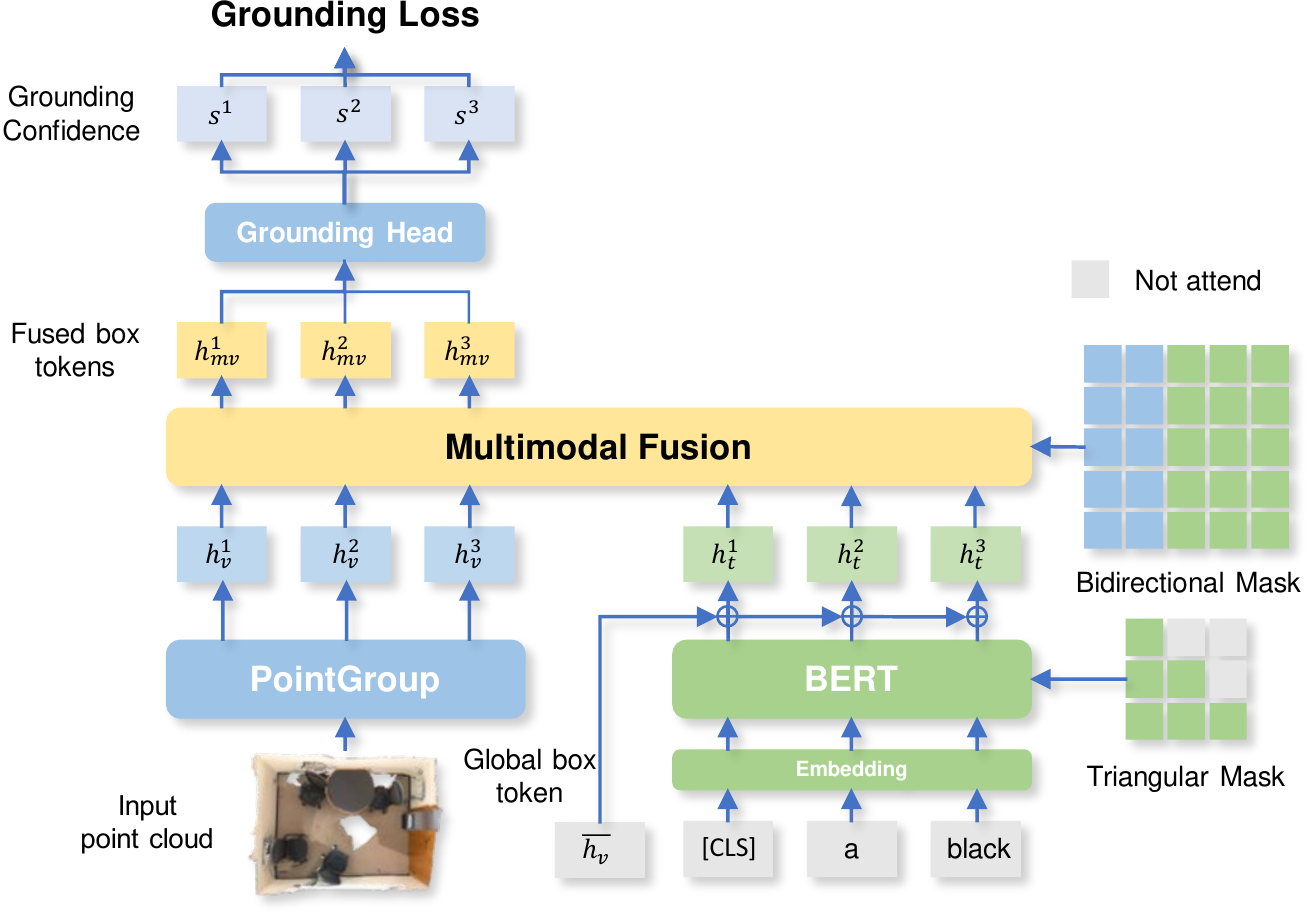}
         \caption{Training flow for 3D visual grounding}
         \label{fig:grounding}
     \end{subfigure}
     \hfill
     \begin{subfigure}[b]{0.48\textwidth}
         \centering
         \includegraphics[width=\textwidth]{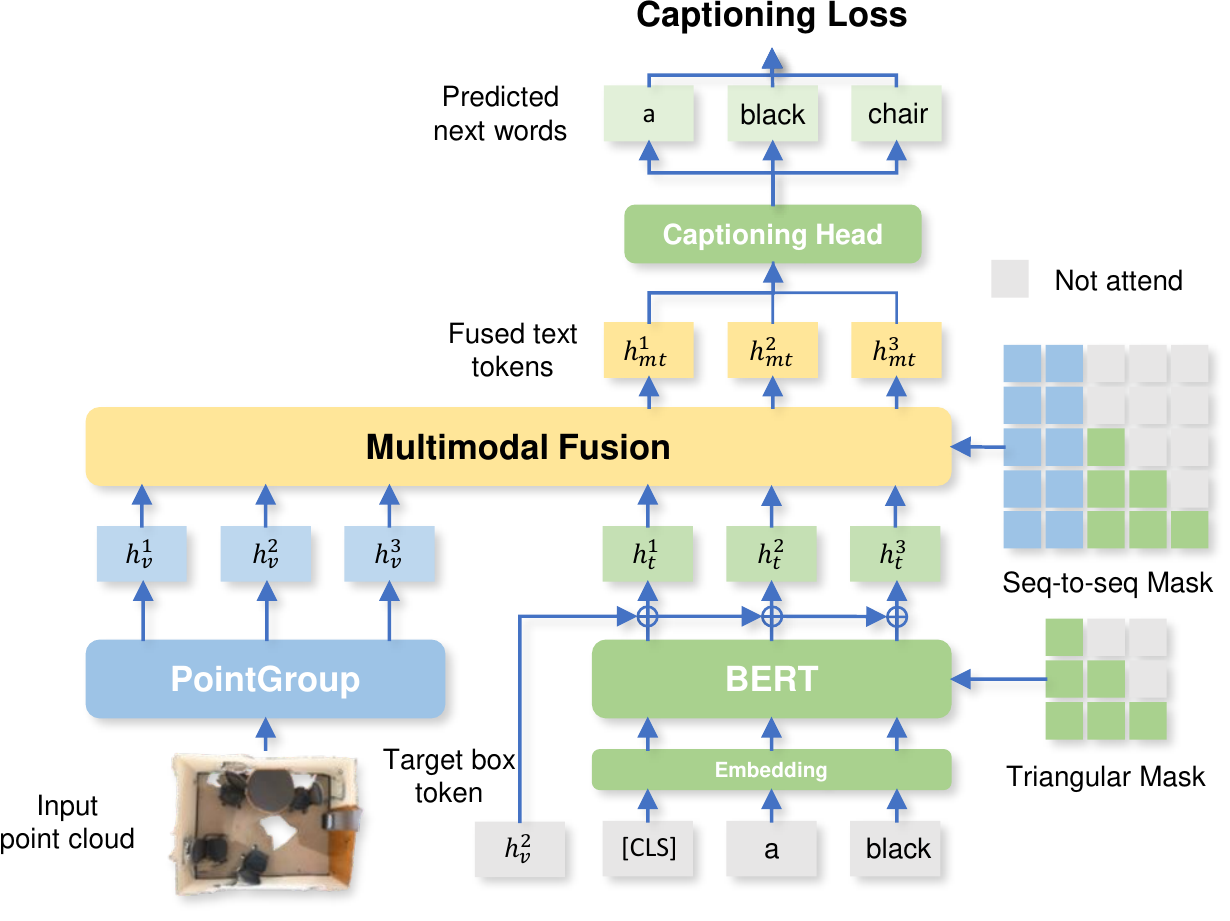}
         \caption{Training flow for 3D dense captioning}
         \label{fig:captioning}
     \end{subfigure}
    \caption{Task-specific training schemes for~\ARCH. The input point cloud is fed into a PointGroup~\citep{jiang2020pointgroup} detection backbone to generate box tokens $\{h_v^i\}$. Parallelly, the text query is processed with a pre-trained BERT~\citep{devlin2018bert} to produce text tokens $\{h_t^i\}$. For visual grounding, as shown in~\ref{fig:grounding}, the multimodal fusion module takes in the concatenated box and text tokens to produce the fused box tokens $\{h^i_{mv}\}$. Then, a lightweight grounding head predicts the grounding confidence scores for each proposal. Similarly, for dense captioning, the current target box token from PointGroup (such as $h_v^2$) is added to all BERT outputs as the captioning cue. The captioning head then takes in the fused text tokens $\{h^i_{mt}\}$ from the multimodal fusion module, and predicts the next tokens in the input sequence, as shown in~\ref{fig:captioning}. Note that a multimodal triangular masking scheme is applied here to enforce the left-to-right sequence modeling.}
    \label{fig:task-specific}
    \vspace{-3mm}
\end{figure*}

%% file: sections/3_method.tex
\section{Method}

Our architecture stem consists of three main opponents as shown in Fig.~\ref{fig:task-specific}: a PointGroup detection backbone, a pre-trained BERT~\citep{devlin2018bert} encoder, and a transformer-based multimodal fusion module. Given a point-cloud-text pair as input, PointGroup takes in the point cloud $\textbf{P} \in \mathcal{R}^{N \times (3+K)}$ to produce a sequence of $M$ box tokens $\{h_v^1,...,h_v^i,...,h_v^M\}$. We follow~\citet{chen2020scanrefer} to construct the auxiliary point features with multi-view features and point normals ($k=134$). In the meantime, the BERT encoder encodes the input text of length $L$ into a sequence of text tokens $\{h_t^1,...,h_t^j,...,h_t^L\}$. We use [CLS] and [SEP] as the start and end tokens of the input text in the BERT encoder, and apply a triangular mask to enforce the left-to-right sequence modeling. We also add a visual context token (from a global box for grounding or a target box for captioning) to every text token $h_t^j$. Finally, the multimodal fusion module enriches the concatenated box-text input sequence and produces a sequence of fused tokens as output.

\subsection{A unified model with task-specific objectives} 

\paragraph{3D visual grounding with bidirectional objective.}
\label{sec:grounding}
 
As shown in Fig.~\ref{fig:grounding}, the input point cloud is processed by PointGroup to produce the instance masks and the box tokens $\{h_{v}^1,...,h_{v}^M\}$. In the meantime, an object text query is fed into BERT to produce the text tokens $\{h_{t}^1,...,h_{t}^L\}$. Here, we average all box tokens as the global box token $\hat{h_{v}}$ and add it to all text tokens as the global visual cue. To be consistent with the captioning task, we constantly apply a triangular mask to enforce the left-to-right modeling. Then, the box tokens are concatenated with the text tokens and fed into the multimodal fusion module to acquire the fused multimodal features. We apply a bidirectional mask in the multimodal fusion module to enable the sequence encoding in both directions. Afterwards, we feed the fused box tokens $\{h_{mv}^1,...,h_{mv}^M\}$ to the lightweight grounding head to predict the grounding confidence scores for each object proposal. The box with the highest grounding confidence will be taken as the grounding output. 
Here, we use a conventional cross-entropy loss $L_{\text{G}}$ for supervision, where the object proposal with the highest IoU with the GT queried box is treated as training GT. Inspired by~\citet{chen2020scanrefer}, we also attach a lightweight MLP taking the encoded [CLS] token as input to predict the object semantic class from the text query. We supervise the language object classification with another cross-entropy loss $L_{\text{cls}}$. 
Since the PointGroup is fine-tuned end-to-end, we apply the original PointGroup loss identical to~\citet{jiang2020pointgroup}, where $L_{\text{PG}}=L_{\text{sem}}+L_{\text{o\_dir}}+L_{\text{o\_reg}}+L_{\text{c\_score}}$. 
Overall, we combine the overall visual grounding loss as: $L_{\text{FG}}=L_{\text{PG}}+L_{\text{G}}+L_{\text{cls}}$.

\vspace{-4mm}
\paragraph{3D dense captioning with seq-to-seq objective.}
\label{sec:captioning}

We follow the next word prediction strategy to train our model, where a lightweight MLP is attached to the task-agnostic multimodal fusion module as the captioning head. As shown in Fig.~\ref{fig:captioning}, the input text is padded with [CLS] and [SEP] token at the beginning and the end of the sentence, respectively. Following the teacher-forcing scheme, for a word sequence $\{w^1,...,w^L\}$ of length $L$, we take the $1^{\text{st}}$ to the $(L-1)^{\text{th}}$ words as input and choose the $2^{\text{nd}}$ to the $L^{\text{th}}$ words as the modeling target. Note that a triangular-style mask is applied to the multimodal fusion module. This way, each text token can only attend to the context box tokens and the other text token before itself in the sequence. 
We apply the seq-to-seq objective loss $L_{\text{C}}$ on the predicted sequence, where a word-level cross-entropy loss is applied on each predicted next word against the target word. 
Similar to Sec.~\ref{sec:grounding}, we also apply the same PointGroup loss $L_{\text{PG}}$ for fine-tuning PointGroup. In the end, we combine the overall dense captioning loss as: $L_{\text{FC}}=L_{\text{PG}}+L_{\text{C}}$.

\vspace{-4mm}
\paragraph{Joint training with both objectives.}

On top of the task-agnostic multimodal fusion module, our architecture can easily enable joint training with both bidirectional and seq-to-seq objectives. In this case, both output heads are attached to the multimodal fusion module. We first input the point cloud and the padded text into PointGroup and BERT to encode the box and text tokens, respectively. To accumulate gradients for both objectives before back-propagation, we apply two forward passes through the multimodal fusion module. In the first pass for the bidirectional objective, the global box token and bidirectional mask are applied. Similarly, the target box token and seq-to-seq mask are applied for the seq-to-seq objective in the second pass. We combine all losses for both objectives as: $L_{\text{FJ}}=L_{\text{PG}}+L_{\text{G}}+L_{\text{cls}}+L_{\text{C}}$.

\subsection{3D pre-training with synthetic data}

\input{figures/synthetic_data}

\paragraph{Synthesize 3D vision-language data from images.}

Since the existing 3D vision-language data~\citep{chen2020scanrefer} is relatively limited in scale and variety, augmenting the training data with vision-language knowledge acquired from existing 2D datasets and models can potentially enable more generic vision-language representation learning for aforementioned 3D vision-language tasks.
As illustrated in Fig.~\ref{fig:synthetic_data}, given a ScanNet frame, we crop out the dominant objects and feed the image crops to CLIPCap~\citep{mokady2021clipcap}, an image captioner empowered by pre-trained CLIP~\citep{radford2021clip} and GPT-2~\citep{radford2019gpt}, to generate its description. Then, we crop out the points in the camera frustum from ScanNet point clouds as the 3D context. We pair the generated description and the cropped point cloud as a synthetic point-cloud pair for this view. To ensure the quality of the created point-cloud-text pairs, we remove pairs whose CLIP similarities between the original image crops and descriptions are lower than 0.3. We crop multiple objects from a single ScanNet frame, and repeatedly perform the description generation and point cloud cropping on every 20 frames in ScanNet. This process produces a large synthetic 3D vision-language dataset for the pre-training purpose.

\input{figures/pre-training}

\vspace{-3mm}
\paragraph{Pre-training on synthetic data with both objectives.}
\label{sec:pre-training}

As illustrated in Fig.~\ref{fig:pre-training}, we feed the point cloud into a PointGroup~\citep{jiang2020pointgroup} detection module to produce a sequence of encoded object tokens. To secure the quality of each object token, the PointGroup module is first trained on the ScanNet instance segmentation task until convergence. During pre-training, we freeze the PointGroup module and use the GT instance masks for encoding the object tokens. In the meantime, the object text query is fed into the BERT~\citep{devlin2018bert} to form a sequence of encoded text tokens. These two sets of tokens are concatenated as a sequence for vision-language modeling. Then, a multimodal transformer module takes the concatenated sequence to produce a sequence of fused multimodal tokens. To uniformly supervise the joint representation learning, we apply bidirectional objective ($L_{\text{G}}+L_{\text{cls}}$) and seq-to-seq objective ($L_{\text{C}}$) introduced in Sec.~\ref{sec:grounding}.
We combine the overall pre-training loss function as: $L_{\text{PT}}=L_{\text{G}}+L_{\text{cls}}+L_{\text{C}}$.

As the synthetic 3D vision-language data contain many noisy samples, after convergence on the synthetic data, we continue the joint training with both objectives on the ScanRefer data with clearer annotations as a second pre-training stage. Then, we separately fine-tune the model on the 3D grounding task and 3D dense captioning task with bidirectional objective and seq-to-seq objective, respectively. We show that this pre-training scheme further boosts the overall accuracy on the downstream tasks. 

\subsection{Inference}

We use task-specific heads for decoding visual grounding and dense captioning outputs from the task-agnostic multimodal fusion module. When inferencing the object captions, we autoregressively generate words from the [CLS] token until the [SEP] token or reaching the maximum length. Following~\citet{chen2021d3net}, we take the minimum and maximum coordinates in the predicted instance masks to construct the object bounding boxes. For the bounding boxes that are assigned to the same GT box, we keep only the box with the highest IoU with the GT box. Note that the bounding boxes used for validating the detection and dense captioning performance are deliberately kept identical.

%% file: figures/synthetic_data.tex
\begin{figure}[!t]
    \centering
    \includegraphics[width=0.99\linewidth]{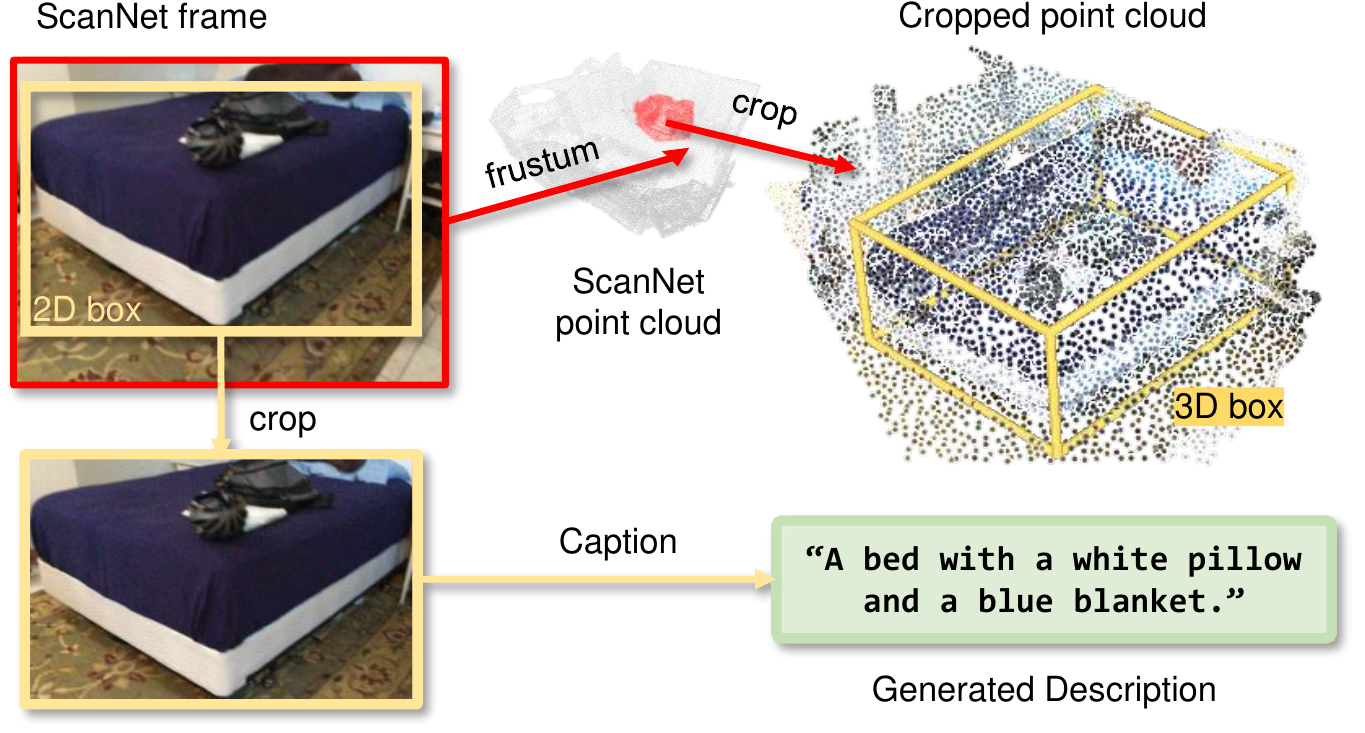}
    \caption{
    We propose a novel vision-language data synthesis method to enable more generic vision-language representation learning for 3D visual grounding and dense captioning,
    In particular, we crop out the dominant objects from the ScanNet frames and feed the image crops to CLIPCap~\citep{mokady2021clipcap}, an off-the-shelf image captioner equipping CLIP~\citep{radford2021clip} and GPT-2~\citep{radford2019gpt}. In the end, we obtain the point-cloud-text pairs by combining the cropped point clouds in the camera frustums and the generated object captions.}
    \label{fig:synthetic_data}
    \vspace{-3mm}
\end{figure}

%% file: figures/pre-training.tex
\begin{figure}[!t]
    \centering
    \includegraphics[width=0.99\linewidth]{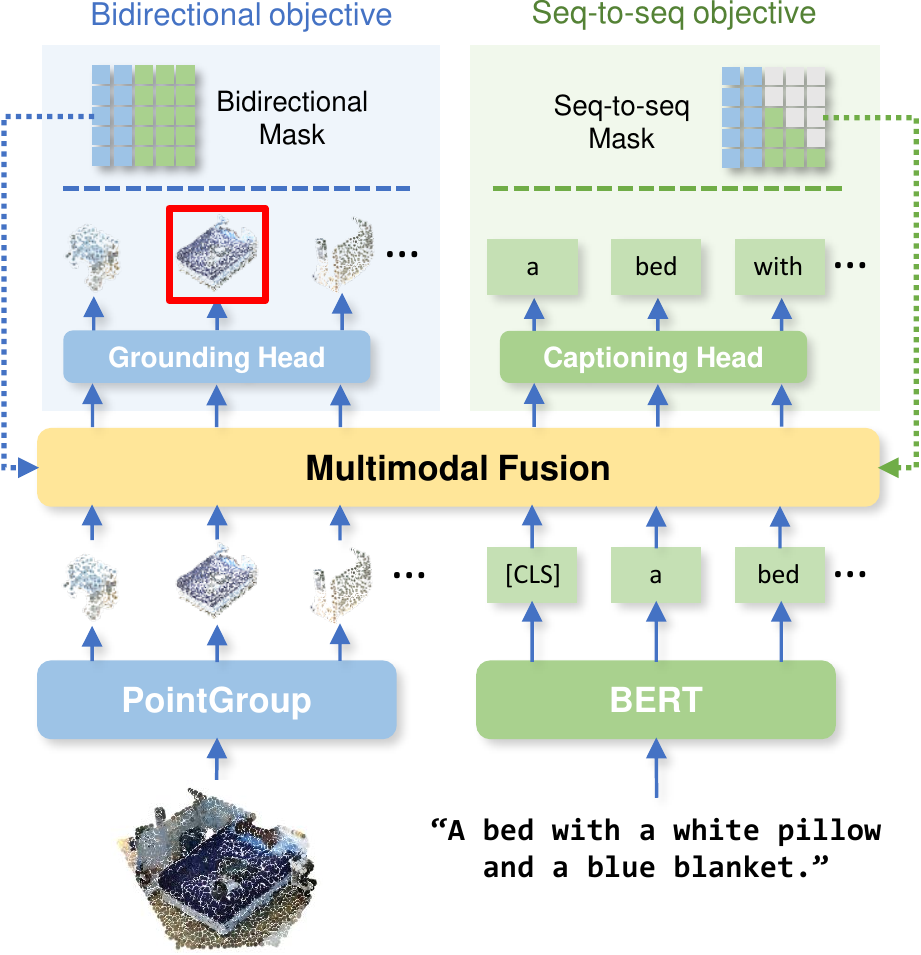}
    \caption{Overview of our 3D vision-language pre-training. A point cloud is fed into a PointGroup~\citep{jiang2020pointgroup} to form a sequence of encoded object tokens. In the meantime, a text query for an object is fed to a BERT~\citep{devlin2018bert} module to form a sequence of encoded text tokens. Based on a lightweight grounding head with a bidirectional mask and a captioning head with a seq-to-seq mask, the objectives here are to predict the queried object from the candidate pool and the next words in the sequence, respectively. We use two simple cross-entropy losses for both objectives.}
    \label{fig:pre-training}
    \vspace{-3mm}
\end{figure}

%% file: sections/4_experiments.tex
\section{Experiments}

\subsection{Implementation details}

Following~\citet{chen2021d3net}, we use PointGroup implemented with the Minkowski Engine~\citep{choy20194d}. The PointGroup backbone is trained on the ScanNet~\citep{dai2017scannet} train set for 500 epochs using the Adam optimizer~\citep{kingma2014adam} with a learning rate of 2e-3 and a batch size of 4. In the pre-training stage on synthetic data, we acquire the object features with a trained and frozen PointGroup. For computational simplicity, we feed the PointGroup with GT instance masks. We consider up to 16 instances in the cropped point clouds. The pre-training takes 40k iterations to converge, with a learning rate of 1e-5 and a batch size of 128. During fine-tuning, we set the learning rate of PointGroup and the rest of the~\ARCH~to 1e-4 and 1e-5, respectively. To boost the fine-tuning speed, we pair each point cloud with 16 descriptions, as in~\citet{chen2021d3net}. Our full architecture contains 120M trainable parameters. All our experiments are conducted on an RTX A6000 GPU with PyTorch~\citep{paszke2019pytorch}.

\subsection{Dataset}

We use the ScanRefer~\citep{chen2020scanrefer} dataset consisting of around 51k descriptions for over 11k objects in 800 ScanNet~\citep{dai2017scannet} scans for the visual grounding and dense captioning tasks.
The descriptions include information about the appearance of the objects, as well as the object-to-object spatial relationships.
We follow the official splits of ScanRefer for training and validation, and report visual grounding and dense captioning results on the validation split.

\input{figures/samples}

% \paragraph{Synthetic data}
\mypara{Synthetic data.}
We repeatedly generate the object descriptions in every 20 frames from ScanNet~\citep{dai2017scannet}. In each frame, we arrange the objects in descending order by the instance mask areas. Up to 3 most dominant objects are cropped for description generation. For consistency, we only generate synthetic data from scans in the ScanNet training split. In the end, we generate $265\,693$ synthetic samples from $92\,766$ frames in $1\,199$ ScanNet~\citep{dai2017scannet} scans. We visualize some synthetic samples in Fig.~\ref{fig:samples}.

\subsection{Evaluation metrics}

\paragraph{Localization.}

Following~\citet{chen2020scanrefer}, we measure the thresholded accuracy Acc@kIoU where the positive predictions have a higher intersection over union (IoU) with the ground truths than the thresholds. We set the threshold value k for IoU to 0.25 and 0.5 in our experiments.

\vspace{-3mm}
\paragraph{Dense captioning.}

To jointly measure the quality of the generated descriptions and the detected bounding boxes, we evaluate them by evaluating standard image captioning metrics such as CIDEr and BLEU under different Intersection-over-Union (IoU) scores between predicted bounding boxes and the matched ground truth bounding boxes. 

For $N^{\text{pred}}$ predicted bounding boxes and $N^{\text{GT}}$ ground truth bounding boxes, we define the captioning precision $M^{\text{P}}@k\text{IoU}$ as $M^{\text{P}}@k\text{IoU} = \frac{1}{N^{\text{pred}}} \sum^{N^{\text{pred}}}_{i=1} m_i u_i$,
% \begin{equation}
%     
% \end{equation}
where $u_i \in \{0, 1\}$ is set to 1 if the IoU score for the $i^{th}$ box is greater than k, otherwise 0. We use $m$ to represent the captioning metrics such as CIDEr, BLEU, METEOR, and ROUGE. 

Similarly, we define the captioning recall $M^{\text{R}}@k\text{IoU}$ as $M^{\text{R}}@k\text{IoU} = \frac{1}{N^{\text{GT}}} \sum^{N^{\text{GT}}}_{i=1} m_i u_i$.
% \begin{equation}
%     M^{\text{R}}@k\text{IoU} = \frac{1}{N^{\text{GT}}} \sum^{N^{\text{GT}}}_{i=1} m_i u_i
% \end{equation}
Note that previous dense captioning metrics proposed in~\citet{chen2021scan2cap} are analogous to $M^{\text{R}}@k\text{IoU}$. It solely measures the caption quality against the matched grounding truth, without taking the false positive predictions into account. 

Finally, we adopt the captioning F1-score combining both captioning precision and captioning recall as the final metric for dense captioning:

\vspace{-3mm}
\begin{equation}
    M@k\text{IoU} = \frac{2 \times M^{\text{P}}@k\text{IoU} \times M^{\text{R}}@k\text{IoU} }{M^{\text{P}}@k\text{IoU} + M^{\text{R}}@k\text{IoU}}
\end{equation}

\vspace{-3mm}
\paragraph{Object detection.}

We also report the mean average precision (mAP) of the detected objects from our PointGroup backbone on ScanRefer val split. These detected boxes are the same as the ones used for dense captioning evaluation.

\subsection{Comparison with the state-of-the-art methods}

We compare our proposed \ARCH~method with several state-of-the-art methods for 3D visual grounding and dense captioning tasks on ScanRefer dataset~\citep{chen2020scanrefer} in Tab.~\ref{tab:grounding} and Tab.~\ref{tab:captioning}. For 3D visual grounding, we divide the validation set into ``Unique'', ``Multiple'', and ``Overall'', following the evaluation protocol in~\citet{chen2020scanrefer}. For 3D dense captioning, we report the aforementioned dense captioning F1-score and the object detection mAP.

\vspace{-3mm}
\paragraph{3D visual grounding.}

\input{tables/grounding}

\input{tables/synthetic_data}

Tab.~\ref{tab:grounding} compares our method with the prior state-of-the-art 3D visual grounding methods on ScanRefer. For our method trained from scratch (Ours (from scratch)), we observe that it already achieves on-par performance with the previous SOTA method~\citep{jain2022bottom}. In this case, our method even achieves the best grounding accuracy in the ``Unique'' subset, where there is only one unique object belonging to a specific semantic class in the scene. After pre-training the network on the synthetic 3D vision-language data and fine-tuning on the visual grounding tasks (Ours (w/ pre-training)), we observe a clear performance boost in the visual grounding accuracies. Despite a drop in the ``Unique'' subset, our method performs clearly better in the more challenging ``Multiple'' subset, resulting in an improvement in the overall grounding accuracy. The improvement in ``Multiple'' subset indicates that pre-training on a large amount of synthetic data provides more multimodal knowledge for disambiguating the objects in the scene with language cues. Note that BUTD-DETR~\citep{jain2022bottom} is re-evaluated by removing the GT object labels in the text queries for a fair comparison (more details in supplementary material).
Besides the modulated object detector, BUTD-DETR also takes pre-computed object bounding boxes as extra visual cues. In contrast, our method follows a much simpler design philosophy and relies purely on the end-to-end fine-tuned object detector. We note that the baseline joint 3DJCG~\citep{cai20223djcg} and D3Net~\citep{chen2021d3net} involves either sophisticatedly designed heavy neural heads or complicated self-critical training strategy. Our method outperforms both of them with a significantly simpler architecture design. 

\vspace{-3mm}
\paragraph{3D dense captioning.}

\input{tables/captioning}

We present a comparison of our method against the previous state-of-the-art 3D dense captioning methods on ScanRefer in Tab.~\ref{tab:captioning}. To keep the comparison consistent and fair, all methods presented here are trained with the cross-entropy objective only, including D3Net~\citep{chen2021d3net}. 
Even without any pre-training, our architecture (Ours (from scratch)) already achieves similar dense captioning results in comparison with the previous SOTA D3Net~\citep{chen2021d3net}. We observe a significant performance boost if we initialize the network with the pre-trained weights on the synthetic data (Ours (w/ pre-training)). We observe that our method clearly outperforms previous joint architectures D3Net and 3DJCG~\citep{cai20223djcg} without any complex design in the architecture or training objectives. Note that D3Net~\citep{chen2021d3net} applies the same PointGroup detection backbone as in our method. Compared to D3Net~\citep{chen2021d3net}, the results show that our method is more capable of modeling the multimodal and spatial relationships from similar visual cues.

\input{tables/synthetic_grounding}

\input{tables/synthetic_captioning}

\subsection{Ablation and analysis}

\paragraph{Does more synthetic data help?}

Since synthetic data can be acquired at almost zero cost in terms of human annotation, there is no theoretical upper bound for the amount of synthetic data we can generate. We conduct an ablation experiment where we pre-train the proposed network on the different amounts of synthetic data. We crop up to 1, 3, or 5 dominant objects from the ScanNet~\citep{dai2017scannet} frames to generate synthetic descriptions. The proposed network is pre-trained on these three synthetic datasets and evaluated on the ScanRefer~\citep{chen2020scanrefer} validation set afterwards. As shown in Tab.~\ref{tab:synthetic_data}, we empirically find that the network pre-trained with the dataset where up to 3 objects are cropped from the frame achieves the best performance for both visual grounding and dense captioning. This indicates that more synthetic data is not necessarily helpful since more noise resulting from it could overwhelm the network in the pre-training stage. 

\vspace{-3mm}
\paragraph{Do pre-training and joint training objectives help?}

\input{figures/line_chart}

We report the ablation results on different pre-training schemes for visual grounding and dense captioning in Tab.~\ref{tab:synthetic_grounding} and Tab.~\ref{tab:synthetic_captioning}, respectively.
In general, we validate 5 scenarios: (a) training the network from scratch on each task in ScanRefer with target objective; (b) training the network from scratch on both tasks in ScanRefer with joint objectives; (c) pre-training the network on synthetic data without further any fine-tuning; (d) fine-tuning the network from (c) directly on the downstream objective; and (e) our full setting -- fine-tuning the network from (c) joint training objectives on ScanRefer. 

For visual grounding in Tab.~\ref{tab:synthetic_grounding}, our model in (a) already demonstrates strong performance on par with the previous SOTA even without pre-training. 
And the visual grounding accuracy is further improved by having the joint training objectives on ScanRefer in (b), showing that our unified architecture effectively connects the two tasks through joint learning. 
After being pre-trained only on the synthetic data, our model in (c) demonstrates non-trivial results when directly evaluated on the ScanRefer~\citep{chen2020scanrefer} validation split. 
Then, the model fine-tuned with only the bidirectional objective in (d) achieves the highest visual grounding accuracy in the ``Unique'' split, indicating a great capability of distinguishing the objects via semantic information.
Fine-tuned jointly with both training objectives, our final model in (e) achieves the highest visual grounding accuracy in the ``Multiple'' subset, leading to the best overall visual grounding results.
Compared with directly training from scratch in (a), this pre-training and joint optimization scheme clearly demonstrates its advantage.

For dense captioning in Tab.~\ref{tab:synthetic_captioning}, our method in (a) already demonstrates competitive performance against the previous SOTA without pre-training, and can be further improved by having the joint training objectives on ScanRefer in (b), indicating the effectiveness of our architecture in unifying two downstream tasks. 
Due to the domain gap, the descriptions generated by the model after pre-training only on the synthetic data in (c) obviously deviate from the GT samples in ScanRefer~\citep{chen2020scanrefer} validation split.
Plausible dense captioning performance is achieved after the network is directly fine-tuned on ScanRefer~\citep{chen2020scanrefer} with the seq-to-seq objective in (d). 
A further performance boost is observed if the pre-trained network is fine-tuned jointly with both the bidirectional and the seq-to-seq objectives in (e).
Here, we also demonstrate the strong advantage of having the pre-training and joint optimization scheme in Fig.~\ref{fig:line_chart}. Compared to training from scratch, our pre-training scheme on synthetic data with joint training objectives effectively improves the overall accuracy on 3D dense captioning and visual grounding.

%% file: figures/samples.tex
\begin{figure*}[!ht]
    \centering
    \includegraphics[width=0.99\linewidth]{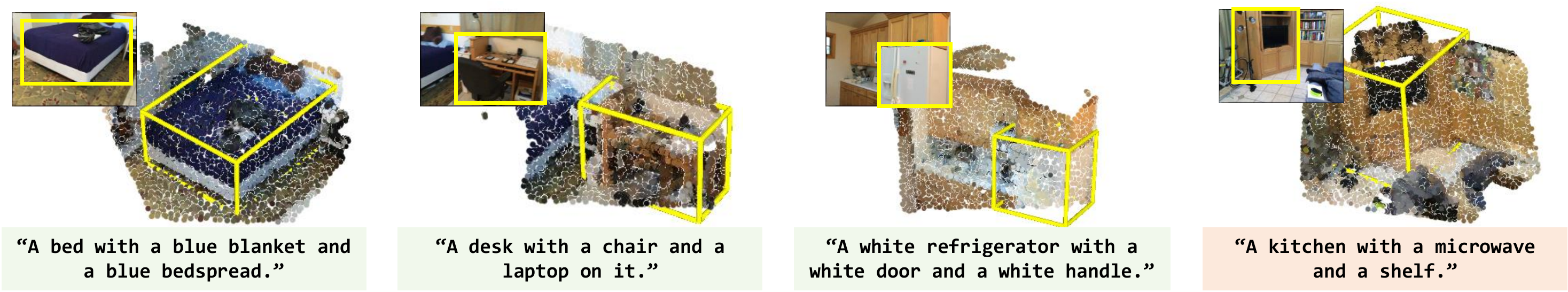}
    \caption{Our generated synthetic samples. Objects in clear views can be well captured by CLIPCap~\citep{mokady2021clipcap} as shown in the first three boxes in green. However, CLIPCap also fails to describe blurry or incomplete objects as shown in the last red box. Figure best viewed in color.}
    \label{fig:samples}
    \vspace{-2mm}
\end{figure*}

%% file: tables/grounding.tex
\begin{table}[!t]
    \centering
    \resizebox{\linewidth}{!}{
        \begin{tabular}{l|ccc|ccc}
            \toprule
            & \multicolumn{3}{c}{Val Acc@0.25IoU} & \multicolumn{3}{c}{Val Acc@0.5IoU} \\ 
            \cmidrule(lr){2-4} \cmidrule(lr){5-7}
            & Unique & Multiple & Overall & Unique & Multiple & Overall \\
            \midrule
            ScanRefer~\citep{chen2020scanrefer} & 76.33 & 32.73 & 41.19 & 53.51 & 21.11 & 27.40 \\
            TGNN~\citep{huang2021text} & 68.61 & 29.84 & 37.37 & 56.80 & 23.18 & 29.70 \\
            InstanceRefer~\citep{yuan2021instancerefer} & 75.72 & 29.41 & 38.40 & 66.83 & 24.77 & 32.93  \\
            3DVG-Trans~\citep{zhao20213dvg} & 81.93 & 39.30 &  47.57 & 60.64 & 28.42 & 34.67 \\
            3DJCG~\citep{cai20223djcg} & 78.75 & 40.13 & 47.62 & 61.30 & 30.08 & 36.14  \\
            D3Net~\citep{chen2021d3net} & - & - & - & 72.04 & 27.11 & 35.58  \\
            3D-SPS~\citep{luo20223dsps} & 84.12 & 40.32 & 48.82 & 66.72 & 29.82 & 36.98 \\
            BUTD-DETR~\citep{jain2022bottom} & 82.77 & \textbf{44.01} & \textbf{49.69} & 63.81 & \textbf{33.51} & 38.01 \\
            \midrule
            Ours (from scratch) & \textbf{85.84} & 32.21 & 42.31 & \textbf{74.78} & 27.60 & 36.49  \\
            Ours (w/ pre-training) & 82.75 & 36.36 & 45.27 & 73.14 & 31.05 & \textbf{39.14}  \\
            \bottomrule
        \end{tabular}
    }
    \vspace{-3mm}
    \caption{Quantitative results on 3D visual grounding. We follow the evaluation setting in~\citet{chen2020scanrefer}. All accuracies are thresholded by the IoU 0.25 and 0.5. In comparison with the previous SOTA methods, our method has a notably higher overall grounding accuracy thresholded by IoU 0.5. Note that the grounding results from BUTD-DETR~\citep{jain2022bottom} are re-evaluated by removing the GT object labels in the text queries from the original implementation.} 
    \label{tab:grounding}
    \vspace{-2mm}
\end{table}

%% file: tables/synthetic_data.tex
\begin{table*}[!t]
    \centering
    \resizebox{\linewidth}{!}{
        \begin{tabular}{l|ccc|cccc}
            \toprule
            & \multicolumn{3}{c}{Visual Grounding Accuracy} & \multicolumn{4}{c}{Dense Captioning F1-Scores} \\
            \# objects & Unique@0.5IoU & Multiple@0.5IoU & Overall@0.5IoU & CIDEr@0.5IoU & BLEU-4@0.5IoU & ROUGE-L@0.5IoU & METEOR@0.5IoU \\
            \midrule
            up to 1 & 26.72 & 7.24 & 11.02 & 3.90 & 0.53 & 18.70 & 8.17 \\
            up to 3 & \textbf{30.84} & \textbf{9.34} & \textbf{13.51} & \textbf{5.91} & \textbf{0.69} & \textbf{18.91} & \textbf{8.32} \\
            up to 5 & 29.54 & 7.62 & 11.87 & 2.81 & 0.29 & 18.22 & 7.89 \\
            \bottomrule
        \end{tabular}
    }
    \vspace{-3mm}
    \caption{Comparison of 3D visual grounding and dense captioning results evaluated on ScanRefer~\citep{chen2020scanrefer} val set with \ARCH~pre-trained on different amount of synthetic data. When using synthetic data generated from up to 3 object crops in the ScanNet~\citep{dai2017scannet} frames, the pre-trained~\ARCH~demonstrates the best initial performance in comparison with other object cropping configurations.}
    \label{tab:synthetic_data} 
    \vspace{-1mm}
\end{table*}

%% file: tables/captioning.tex
\begin{table}[!t]
    \centering
    \resizebox{\linewidth}{!}{
        \begin{tabu}{l|cccc|c}
            \toprule
            & \multicolumn{4}{c}{Captioning F1-score} & Detection \\
            & C@0.5IoU & B-4@0.5IoU & R@0.5IoU & M@0.5IoU & mAP@0.5 \\
            \midrule
            Scan2Cap~\citep{chen2021scan2cap} & 15.71 & 9.01 & 14.92 & 7.18 & 32.09 \\
            X-Trans2Cap~\citep{yuan2022x} & 17.64 & 9.68 & 15.25 & 7.21 & 35.31 \\
            MORE~\citep{jiao2022more} & 16.46 & 8.86 & 14.71 & 7.12 & 31.93 \\
            3DJCG~\citep{cai20223djcg} & 21.17 & 14.18 & 19.49 & 10.19 & 39.75 \\
            D3Net~\citep{chen2021scan2cap} & 26.13 & 16.18 & 27.48 & 13.06 & 50.93 \\
            \rowfont{\color{lightgray}}
            D3Net~\citep{chen2021scan2cap} (CIDEr loss) & 41.32 & 22.75 & 35.30 & 15.87 & 53.85 \\
            \midrule
            Ours (from scratch) & 26.68 & 14.64 & 27.10 & 12.92 & 53.91 \\
            Ours (w/ pre-training) & \textbf{30.28} & \textbf{18.23} & \textbf{30.72} & \textbf{14.74} & \textbf{54.03} \\
            \bottomrule
        \end{tabu}
    }
    \vspace{-3mm}
    \caption{Quantitative results of 3D dense captioning on ScanRefer~\citep{chen2020scanrefer}. We measure the dense captioning F1-scores and the PointGroup detection mAP against the ground truth bounding boxes and descriptions. All reported metrics are thresholded by IoU 0.5. Our method outperforms the previous SOTA methods even without any pre-training, and our proposed pre-training scheme on the synthetic data gives a further improvement the dense captioning performance. Note that we compare to D3Net~\citep{chen2021d3net} trained only with the cross-entropy objective for a fair comparison.}
    \label{tab:captioning}
    \vspace{-3mm}
\end{table}

%% file: tables/synthetic_grounding.tex
\begin{table*}[!t]
    \centering
    \resizebox{\linewidth}{!}{
        \begin{tabular}{l|cc|cc|ccc}
            \toprule
            & \multicolumn{2}{c}{Training Dataset(s)} & \multicolumn{2}{c}{Training Objective(s)} & \multicolumn{3}{c}{Visual Grounding Accuracy} \\
            Training setup & Synthetic & ScanRefer & Bidirectional & Seq-to-Seq & Unique@0.5IoU & Multiple@0.5IoU & Overall@0.5IoU \\
            \midrule
            (a) direct from scratch &  & \checkmark & \checkmark & & 74.78 & 27.60 & 36.49 \\
            \midrule
            (b) joint from scratch &  & \checkmark & \checkmark & \checkmark & 73.68 & 28.84 & 37.45 \\
            \midrule
            (c) initial pre-trained & \checkmark &  & \checkmark & \checkmark & 30.84 & 9.34 & 13.51 \\
            (d) direct fine-tuned &  & \checkmark & \checkmark &  & \textbf{75.51} & 29.63 & 38.45 \\
            (e) joint fine-tuned &  & \checkmark & \checkmark & \checkmark & 73.14 & \textbf{31.05} & \textbf{39.14} \\
            \bottomrule
        \end{tabular}
    }
    \vspace{-3mm}
    \caption{
    3D visual grounding results on ScanRefer~\citep{chen2020scanrefer} with different training schemes. (a) Without any pre-training, our model already has a strong performance on par with the previous SOTA. (b) The visual grounding accuracy is further improved by having the joint training objectives, indicating the effectiveness of our architecture in unifying two downstream tasks. (c) Pre-trained on the synthetic data alone gives non-trivial visual grounding performance on the ScanRefer~\citep{chen2020scanrefer} validation split. (d) Fine-tuned with bidirectional objective from weights in (c), it achieves the highest visual grounding accuracy in the ``Unique'' subset. (e) Fine-tuned jointly with both objectives, our final setting in achieves the highest visual grounding accuracy in the ``Multiple'' subset, leading to the best overall visual grounding results.}
    \label{tab:synthetic_grounding}
    \vspace{-1mm}
\end{table*}

%% file: tables/synthetic_captioning.tex
\begin{table*}[!t]
    \centering
    \resizebox{\linewidth}{!}{
        \begin{tabular}{l|cc|cc|cccc}
            \toprule
            & \multicolumn{2}{c}{Training Dataset(s)} & \multicolumn{2}{c}{Training Objective(s)} & \multicolumn{4}{c}{Dense Captioning F1-Scores} \\
            Training setup & Synthetic & ScanRefer & Bidirectional & Seq-to-Seq & CIDEr@0.5IoU & BLEU-4@0.5IoU & ROUGE-L@0.5IoU & METEOR@0.5IoU \\
            \midrule
            (a) direct from scratch &  & \checkmark & \checkmark & & 26.68 & 14.64 & 27.10 & 12.92 \\
            \midrule
            (b) joint from scratch &  & \checkmark & \checkmark & \checkmark & 27.28 & 17.22 & 29.12 & 13.74 \\
            \midrule
            (c) initial pre-trained & \checkmark &  & \checkmark & \checkmark & 5.91 & 0.69 & 18.91 & 8.32 \\
            (d) direct fine-tuned &  & \checkmark &  & \checkmark & 28.13 & 17.69 & 30.33 & 14.30 \\
            (e) joint fine-tuned & & \checkmark & \checkmark & \checkmark & \textbf{30.28} & \textbf{18.23} & \textbf{30.72} & \textbf{14.74} \\
            \bottomrule
        \end{tabular}
    }
    \vspace{-3mm}
    \caption{3D dense captioning results on ScanRefer~\citep{chen2020scanrefer} with different training schemes. (a) Our model without pre-training already demonstrates competitive performance against the previous SOTA. (b) The dense captioning accuracy is further improved by having the joint training objectives, indicating the effectiveness of our architecture in unifying two downstream tasks. (c) Due to the domain gap, the descriptions generated by the model pre-trained only on the synthetic data obviously deviate from the GT samples in ScanRefer~\citep{chen2020scanrefer} validation split. (d) Plausible dense captioning performance is achieved after being directly fine-tuned on ScanRefer~\citep{chen2020scanrefer} with the seq-to-seq objective. (e) A further performance boost is observed if the network is fine-tuned jointly with both training objectives in our final setting.}
    \label{tab:synthetic_captioning}
    \vspace{-3mm}
\end{table*}

%% file: figures/line_chart.tex
\begin{figure}[!t]
    \centering
    \includegraphics[width=0.99\linewidth]{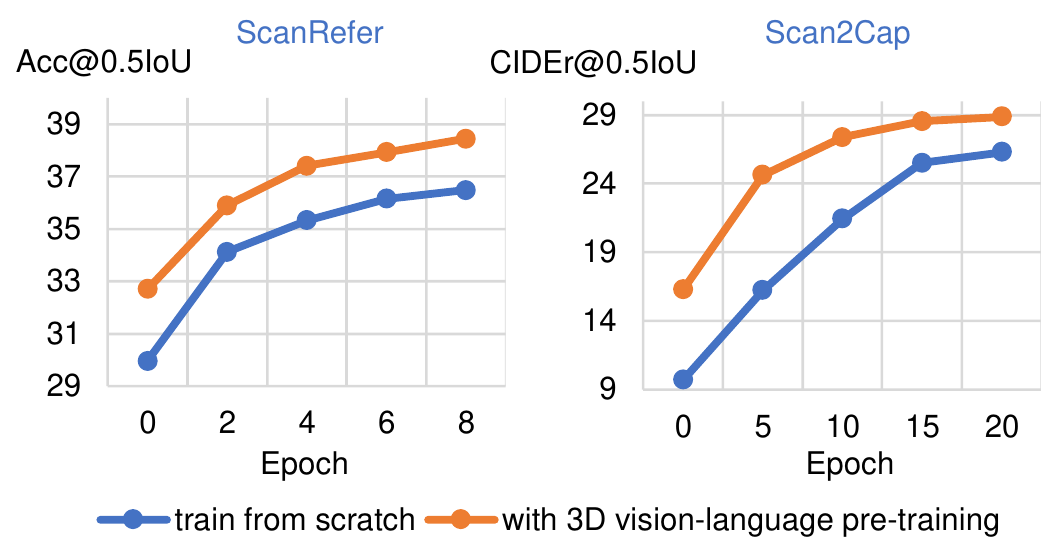}
    \caption{Compared to training from scratch, the pre-training scheme on synthetic data with joint training objectives shows notable benefits on both 3D dense captioning and visual grounding.}
    \label{fig:line_chart}
    \vspace{-3mm}
\end{figure}

%% file: sections/5_conclusion.tex
\section{Conclusion}

We present~\ARCH, a unified transformer architecture to connect 3D dense captioning and visual grounding. 
\ARCH enables learning a strong joint multimodal representation across two tasks through a supervised joint pre-training scheme with bidirectional and seq-to-seq objectives.
The generic representation of~\ARCH expands pre-training scope to more various training sources such as the synthesized data via 2D priors, showing that the distillate 2D knowledge is beneficial to 3D vision-language tasks.
Extensive experiments and analysis demonstrate the strength of our \ARCH model for 3D dense captioning and visual grounding.
We hope our work can inspire more future work in exploring the 3D vision-language field.

\mypara{Limitations.} 
Although~\ARCH takes a promising step towards unifying the two discussed tasks, ~\ARCH could be still expanded to other tasks. 
Further, in addition to generating synthetic data from a 2D captioner, more sources of distillate 2D data could be explored in the future.

%% file: sections/7_acknowledgements.tex
\section*{Acknowledgements}

Dave Zhenyu Chen and Matthias Nie{\ss}ner's work is funded by Google (AugmentedPerception), the ERC Starting Grant Scan2CAD (804724), and a Google Faculty Award. We would also like to thank the support of the TUM-IAS Rudolf M{\"o}{\ss}bauer and Hans Fischer Fellowships (Focus Group Visual Computing), as well as the the German Research Foundation (DFG) under the Grant \textit{Making Machine Learning on Static and Dynamic 3D Data Practical}.  This work is also supported in part by the Canada CIFAR AI Chair program and an NSERC Discovery Grant.

%% file: sections/6_supplemental.tex
\clearpage
\appendix
\counterwithin{figure}{section}
\counterwithin{table}{section}
\counterwithin{equation}{section}

\section*{Supplementary Material}

In this supplementary material, we provide detailed dense captioning results on the ScanRefer dataset in Sec.~\ref{sec:detailed_captioning}.
To showcase the effectiveness of the proposed pre-training scheme and joint training objectives, we provide additional results and analysis in Sec.~\ref{sec:further_training}.
We also include details about a re-evaluation of BUTD-DETR~\citep{jain2022bottom} in Sec.~\ref{sec:re_evaluation}.

\section{Detailed dense captioning results}
\label{sec:detailed_captioning}

\input{tables/detailed_captioning.tex}

\input{figures/dense_captions.tex}

We present the detailed dense captioning precisions and recalls in Tab.~\ref{tab:detailed_captioning}. To keep the comparison consistent and fair, all methods presented here are trained with the cross-entropy objective only, including D3Net~\citep{chen2021d3net}. As discussed in the Experiments Section in the main paper, the previous evaluation protocol in~\citet{chen2021scan2cap} mainly covers the dense captioning recalls without penalizing false positives. This protocol only takes the number of GT boxes into account, neglecting the fact that up to infinite predictions can be produced without being punished. We further detailed the dense captioning precisions and recalls, as displayed in Tab.~\ref{tab:captioning_precisions} and Tab.~\ref{tab:captioning_recalls}. For some previous methods with VoteNet~\citep{qi2019deep} backbone such as MORE~\citep{jiao2022more} and 3DJCG~\citep{cai20223djcg}, their dense captioning precisions are notably lower than the other methods with stronger detection backbone such as D3Net~\citep{chen2021d3net}. To further showcase the impact of having cleaner box predictions, we visualize the predicted boxes with captions in Fig.~\ref{fig:dense_captions}. Despite having a slightly lower dense captioning recall, our method still produces much more plausible bounding box predictions, resulting in a strong dense captioning precision and F1-score compared with the previous methods.

\section{Further training details and analysis}
\label{sec:further_training}

\input{tables/detailed_synthetic_grounding.tex}

\input{tables/detailed_synthetic_captioning.tex}

As the synthetic data contain many noise samples, we continue the joint training scheme with the bidirectional and seq-to-seq objectives after the convergence on the synthetic data. Additionally, to make sure the multimodal representation contains task-specific information, we further fine-tune the network with the training objective of the specific target task (\ie bidirectional for grounding and seq-to-seq for captioning) on ScanRefer as the final training stage. To show the effectiveness of the joint training objective, we report the intermediate training steps for ``joint from scratch'' and ``joint fine-tuned'' in Sec. 4.5 of the main paper.

In particular, for ``joint from scratch'', we follow a two-stage training strategy. We first train the network from scratch on ScanRefer with both bidirectional and seq-to-seq objectives (a), then continue training the network on ScanRefer with the target objective (b). As shown in Tab.~\ref{tab:detailed_synthetic_grounding} and Tab.~\ref{tab:detailed_synthetic_captioning}, such two-stage \textit{joint-to-target} training strategy can effectively improve both visual grounding accuracy and dense captioning results. These improvements indicate that our network is capable of learning and sharing a strong joint representation across two downstream tasks.

Similarly, to show the advantage of pre-training on the distillate 2D priors, we report the intermediate results of the two-stage training steps for ``joint fine-tuned''. We first train the network from scratch on the synthesized data with both bidirectional and seq-to-seq objectives (c), then fine-tune the pre-trained network on the downstream tasks with the respective target objective (d). By comparing (c) with (a), we observe a clear performance boost in both downstream tasks. Further improvements can be observed after the final fine-tuning step on the downstream task. Such improvements further validate the effectiveness of expanding the multimodal representation learning to distillate 2D data.

\section{Re-evaluation of BUTD-DETR}
\label{sec:re_evaluation}

\input{tables/butd_detr.tex}

We notice that the input text queries of BUTD-DETR~\citep{jain2022bottom} in the official implementation differ from the evaluation protocol in the other work~\citep{chen2020scanrefer, huang2021text, yuan2021instancerefer, zhao20213dvg, cai20223djcg, chen2021d3net}, where the GT object labels are manually added to the text. For instance, given a query for a table ``this is a round wooden object. it is between two black chairs.'', the official implementation adds the GT object label ``table'' to the query as ``this is a round wooden object. \underline{table}. it is between two black chairs.''. Such augmentation during evaluation leads to three problems: 1) Using the GT object labels during inference results in unfair comparison; 2) The rich relationships in the language cues are neglected, as the grounding model tends to rely on the object names to distinguish objects in the scene; 3) Some difficult cases are aided by exposed GT information where the target is simply referred as ``object'' in the query, as in the aforementioned example. For the purpose of having a fair and consistent comparison, we re-evaluate BUTD-DETR~\citep{jain2022bottom} by removing the additional object names in the input texts. The re-evaluated visual grounding results against the ones in the original paper~\citep{jain2022bottom} are displayed in Tab.~\ref{tab:butd_detr}.

%% file: tables/detailed_captioning.tex
\begin{table}[!h]
    \begin{subtable}[h]{0.99\linewidth}
        \centering
        \resizebox{\textwidth}{!}{
        \begin{tabu}{l|cccc|c}
            \toprule
            & \multicolumn{4}{c}{Captioning Precisions} & Detection \\
            & C@0.5IoU & B-4@0.5IoU & R@0.5IoU & M@0.5IoU & mAP@0.5 \\
            \midrule
            Scan2Cap~\citep{chen2021scan2cap} & 10.21 & 5.85 & 9.70 & 4.67 & 32.09 \\
            X-Trans2Cap~\citep{yuan2022x} & 11.04 & 6.00 & 11.54 & 3.92 & 35.31 \\
            MORE~\citep{jiao2022more} & 10.30 & 5.49 & 11.14 & 3.87 & 33.75 \\
            3DJCG~\citep{cai20223djcg} & 13.47 & 9.19 & 16.31 & 5.66 & 39.75 \\
            D3Net~\citep{chen2021scan2cap} & 18.24 & 11.04 & 31.53 & 7.47 & 50.93 \\
            \rowfont{\color{lightgray}}
            D3Net~\citep{chen2021scan2cap} (CIDEr loss) & 30.83 & 16.70 & 26.24 & 11.48 & 53.85 \\
            \midrule
            Ours (from scratch) & 19.92 & 10.25 & 19.43 & 9.28 & 53.91 \\
            Ours (w/ pre-training) & \textbf{22.41} & \textbf{13.70} & \textbf{23.07} & \textbf{11.11} & \textbf{54.03} \\
            \bottomrule
        \end{tabu}
        }
       \caption{3D Dense Captioning Precisions}
       \label{tab:captioning_precisions}
    \end{subtable}
    \hfill
    \begin{subtable}[h]{0.99\linewidth}
        \centering
        \resizebox{\textwidth}{!}{
        \begin{tabu}{l|cccc|c}
            \toprule
            & \multicolumn{4}{c}{Captioning Recalls} & Detection \\
            & C@0.5IoU & B-4@0.5IoU & R@0.5IoU & M@0.5IoU & mAP@0.5 \\
            \midrule
            Scan2Cap~\citep{chen2021scan2cap} & 39.08 & 23.32 & 44.48 & 21.97 & 32.09 \\
            X-Trans2Cap~\citep{yuan2022x} & 43.87 & 25.05 & 44.97 & 22.46 & 35.31 \\
            MORE~\citep{jiao2022more} & 40.94 & 22.93 & 44.42 & 21.66 & 33.75 \\
            3DJCG~\citep{cai20223djcg} & \textbf{49.48} & \textbf{31.03} & 50.80 & 24.22 & 39.75 \\
            D3Net~\citep{chen2021scan2cap} & 46.07 & 30.29 & \textbf{51.67} & \textbf{24.35} & 50.93 \\
            \rowfont{\color{lightgray}}
            D3Net~\citep{chen2021scan2cap} (CIDEr loss) & 62.64 & 35.68 & 53.90 & 25.72 & 53.85 \\
            \midrule
            Ours (from scratch) & 40.40 & 25.60 & 44.75 & 21.26 & 53.91 \\
            Ours (w/ pre-training) & 46.69 & 27.22 & 45.98 & 21.91 & \textbf{54.03} \\
            \bottomrule
        \end{tabu}
    }
        \caption{3D Dense Captioning Recalls}
        \label{tab:captioning_recalls}
     \end{subtable}

     \caption{The 3D dense captioning precisions and recalls on Scan2Cap~\citep{chen2021scan2cap} validation set. All reported metrics are thresholded by IoU 0.5. Our method achieves strong dense captioning precisions and competitive dense captioning recalls in comparison to previous methods. Note that we compare to D3Net~\citep{chen2021d3net} trained only with the cross-entropy objective for a fair comparison.}
     \label{tab:detailed_captioning}
\end{table}

%% file: figures/dense_captions.tex
\begin{figure*}[!h]
    \centering
    \includegraphics[width=0.99\linewidth]{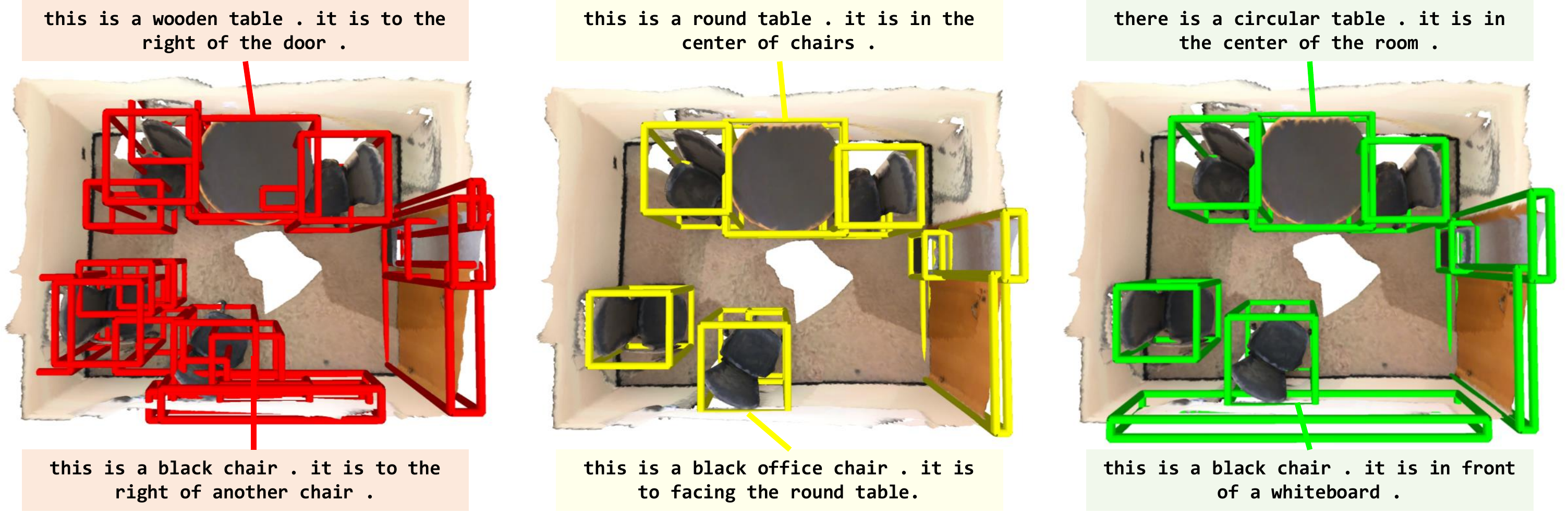}
    \caption{
    Detected boxes with captions from 3DJCG~\citep{cai20223djcg} (in red boxes), our method (in yellow boxes), and ground truths (in green boxes). Our method generates much fewer and cleaner box predictions when compared to those from 3DJCG. This results in a much higher dense captioning precision. Best viewed in color.
    }
    \label{fig:dense_captions}
\end{figure*}

%% file: tables/detailed_synthetic_grounding.tex
\begin{table*}[!t]
    \centering
    \resizebox{\linewidth}{!}{
        \begin{tabular}{l|cc|cc|ccc}
            \toprule
            & \multicolumn{2}{c}{Training Dataset(s)} & \multicolumn{2}{c}{Training Objective(s)} & \multicolumn{3}{c}{Visual Grounding Accuracy} \\
            Training setup & Synthetic & ScanRefer & Bidirectional & Seq-to-Seq & Unique@0.5IoU & Multiple@0.5IoU & Overall@0.5IoU \\
            \midrule
            (a) joint from scratch &  & \checkmark & \checkmark & \checkmark & 72.30 & 28.41 & 36.85 \\
            (b) continue from (a) &  & \checkmark & \checkmark &  & \textbf{73.68} & 28.84 & 37.45 \\
            \midrule
            (c) joint from pre-trained & \checkmark & \checkmark & \checkmark & \checkmark & 72.63 & 30.67 & 38.81 \\
            (d) continue from (c) &  & \checkmark & \checkmark &  & 73.14 & \textbf{31.05} & \textbf{39.14} \\
            \bottomrule
        \end{tabular}
    }

    \caption{
    3D visual grounding results on ScanRefer~\citep{chen2020scanrefer} with detailed pre-training and joint training steps. 
    (a) When trained from scratch on ScanRefer~\citep{chen2020scanrefer} with joint training objectives, our model already has a strong performance on par with the previous SOTA. 
    (b) Continuing fine-tuning from (a) solely with the bidirectional objective improves the visual grounding results. 
    (c) Jointly training with both objectives from pre-trained weights on the synthetic data, it achieves better visual grounding results in comparison with jointly training from scratch (a). 
    (d) Continuing fine-tuning from (c) with the bidirectional objective, our final setting achieves the best overall visual grounding results.
    }
    \label{tab:detailed_synthetic_grounding}
\end{table*}

%% file: tables/detailed_synthetic_captioning.tex
\begin{table*}[!t]
    \centering
    \resizebox{\linewidth}{!}{
        \begin{tabular}{l|cc|cc|cccc}
            \toprule
            & \multicolumn{2}{c}{Training Dataset(s)} & \multicolumn{2}{c}{Training Objective(s)} & \multicolumn{4}{c}{Dense Captioning F1-Scores} \\
            Training setup & Synthetic & ScanRefer & Bidirectional & Seq-to-Seq & CIDEr@0.5IoU & BLEU-4@0.5IoU & ROUGE-L@0.5IoU & METEOR@0.5IoU \\
            \midrule
            (a) joint from scratch &  & \checkmark & \checkmark & \checkmark & 26.48 & 14.64 & 27.10 & 12.92 \\
            (b) continue from (a) &  & \checkmark &  & \checkmark & 27.28 & 17.22 & 29.12 & 13.74\\
            \midrule
            (c) joint from pre-trained & \checkmark & \checkmark & \checkmark & \checkmark & 29.77 & 17.78 & 30.10 & 14.28 \\
            (d) continue from (c) & & \checkmark &  & \checkmark & \textbf{30.28} & \textbf{18.23} & \textbf{30.72} & \textbf{14.74} \\
            \bottomrule
        \end{tabular}
    }

    \caption{
    3D dense captioning results on ScanRefer~\citep{chen2020scanrefer} with detailed pre-training and joint training steps. 
    (a) Our model without pre-training already demonstrates competitive performance against the previous SOTA. 
    (b) Continuing fine-tuning from (a) solely with the seq-to-seq objective improves the dense captioning results. 
    (c) Jointly training the network with pre-trained weights on synthetic data, it achieves better dense captioning results in comparison with jointly training from scratch (a). 
    (d) Continuing fine-tuning from (c) with the seq-to-seq objective, our final setting achieves the best overall dense captioning results.
    }
    \label{tab:detailed_synthetic_captioning}
\end{table*}

%% file: tables/butd_detr.tex
\begin{table}[!t]
    \centering
    \resizebox{\linewidth}{!}{
        \begin{tabular}{l|ccc|ccc}
            \toprule
            & \multicolumn{3}{c}{Val Acc@0.25IoU} & \multicolumn{3}{c}{Val Acc@0.5IoU} \\ 
            \cmidrule(lr){2-4} \cmidrule(lr){5-7}
            & Unique & Multiple & Overall & Unique & Multiple & Overall \\
            \midrule
            Original & 84.20 & 46.60 & 52.20 & 66.30 & 35.10 & 39.80 \\
            Re-evaluated & 82.77 & 44.01 & 49.69 & 63.81 & 33.51 & 38.01 \\
            \bottomrule
        \end{tabular}
    }

    \caption{3D visual grounding accuracy of BUTD-DETR~\citep{jain2022bottom}. We re-evaluated BUTD-DETR~\citep{jain2022bottom} by removing the GT object labels in the text queries from the original implementation.}
    \label{tab:butd_detr}
\end{table}